\documentclass[conference]{IEEEtran}
\usepackage{times}

\usepackage[numbers]{natbib}
\usepackage{multicol}
\usepackage[bookmarks=true]{hyperref}
\usepackage{xcolor}
\usepackage{amsmath}
\usepackage{multirow}
\usepackage{graphicx}
\usepackage{booktabs}
\usepackage{etoolbox}
\makeatletter
\patchcmd{\@makecaption}
  {\scshape}
  {}
  {}
  {}
\makeatletter
\patchcmd{\@makecaption}
  {\\}
  {.\ }
  {}
  {}
\makeatother
\def\tablename{\textbf{Table}}

\newcommand{\TightenPar}[1]{\looseness=-#1}

\hypersetup{colorlinks=true,
            citecolor=[rgb]{0.69,0.16,0.16},
            linkcolor=[rgb]{0.69,0.16,0.16},
            urlcolor=[rgb]{0.69,0.16,0.16}}

\renewcommand{\figurename}{\textbf{Figure}}

\begin{document}

\title{\textit{emg2tendon}: From sEMG Signals to Tendon Control in Musculoskeletal Hands}


\author{\authorblockN{Sagar Verma}
\authorblockA{Independent Researcher\\
Cambridge, MA 02138\\
Email: sagar15056@iiitd.ac.in}
}


%

\thispagestyle{empty}
\pagestyle{empty}

\makeatletter
\let\@oldmaketitle\@maketitle 
\renewcommand{\@maketitle}{\@oldmaketitle 
  \begin{center}
      \includegraphics[width=\linewidth]{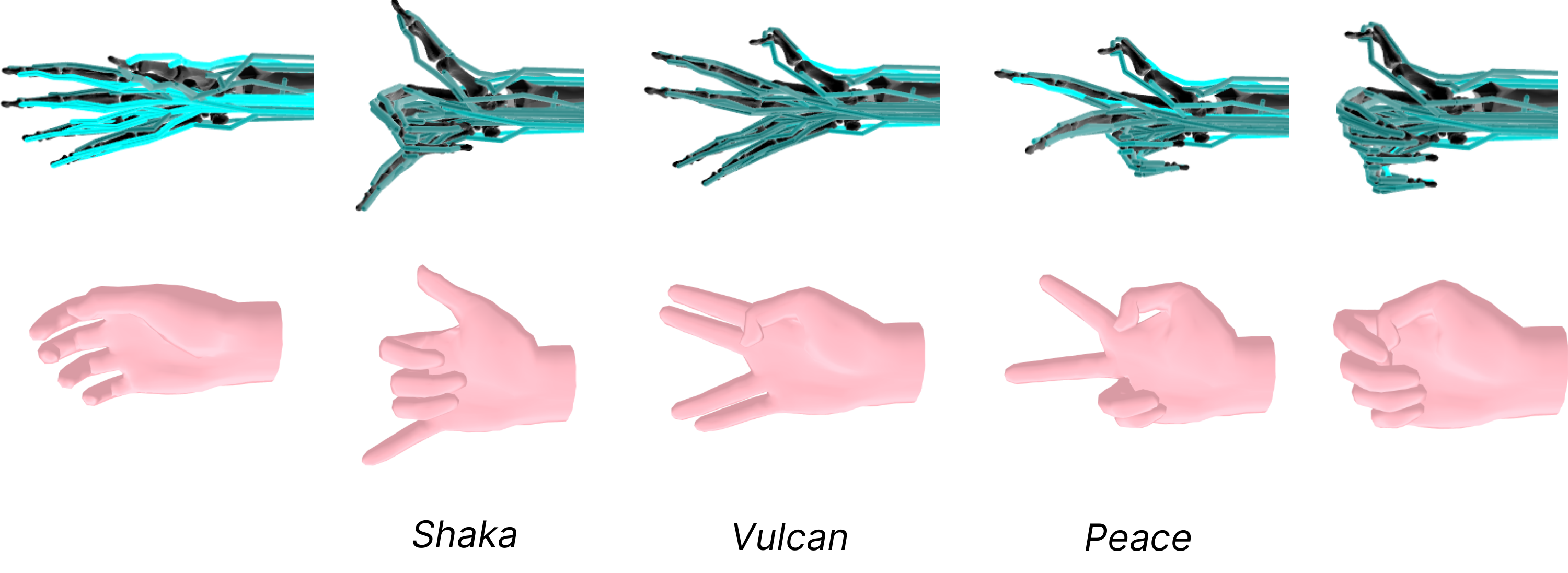}
  \end{center}
  \textbf{Figure }1. We introduce the \textit{\textit{emg2tendon}} dataset, an extension of \textit{\textit{emg2pose}}, for learning tendon control signals from sEMG signals. The bottom row presents hand poses for the gestures ('Shaka,' 'Vulcan,' and 'Peace') sourced from the \textit{\textit{emg2pose}} dataset, while the top row depicts the corresponding musculoskeletal hand poses generated using the proposed pipeline. \bigskip
}
\makeatother

\maketitle

\setcounter{figure}{1}

\begin{abstract}
Tendon-driven robotic hands offer unparalleled dexterity for manipulation tasks, but learning control policies for such systems presents unique challenges. Unlike joint-actuated robotic hands, tendon-driven systems lack a direct one-to-one mapping between motion capture (mocap) data and tendon controls, making the learning process complex and expensive. Additionally, visual tracking methods for real-world applications are prone to occlusions and inaccuracies, further complicating joint tracking. Wrist-wearable surface electromyography (sEMG) sensors present an inexpensive, robust alternative to capture hand motion. However, mapping sEMG signals to tendon control remains a significant challenge despite the availability of EMG-to-pose data sets and regression-based models in the existing literature. 

We introduce the first large-scale EMG-to-Tendon Control dataset for robotic hands, extending the \emph{emg2pose} dataset, which includes recordings from 193 subjects, spanning 370 hours and 29 stages with diverse gestures. This dataset incorporates tendon control signals derived using the MyoSuite MyoHand model, addressing limitations such as invalid poses in prior methods. We provide three baseline regression models to demonstrate \emph{emg2tendon} utility and propose a novel diffusion-based regression model for predicting tendon control from sEMG recordings. This dataset and modeling framework marks a significant step forward for tendon-driven dexterous robotic manipulation, laying the groundwork for scalable and accurate tendon control in robotic hands. \href{https://emg2tendon.github.io/}{https://emg2tendon.github.io/}
\end{abstract}

\IEEEpeerreviewmaketitle

\section{Introduction} \label{sec:intro}

The human hand, with its exceptional dexterity and compliance, serves as an ideal model for robotic hands designed for complex manipulation tasks. From assembling small components in manufacturing to performing intricate surgical procedures, fully anthropomorphic, tendon-driven robotic hands offer the highest level of control and adaptability \cite{Caggiano2023MyoDexAG}. However, traditional visual pose-tracking methods, such as motion capture and depth sensors \cite{Carreras2012IncreasingHF, Darvish2023TeleoperationOH, Wang2020RGB2Hands}, suffer from occlusions, limited field of view, and environmental constraints, making them impractical for continuous real-world use. Wrist-worn surface electromyography (sEMG) sensors provide a viable alternative by capturing neuromuscular signals that drive hand movement \cite{Salter2024emg2poseAL}. Mapping sEMG to tendon control signals enables natural and responsive robotic hand operation, with significant implications for prosthetics, teleoperation, and human-robot interaction \cite{agamemnon2017pros}.

Anthropomorphic hands require intricate joint mechanisms, including hinges, linkages, and tendon-driven actuators, to achieve human-like movement \cite{shadow2023, Puhlmann2022RBOH3, wonik2023, Shaw2023LEAPHL}. Unlike non-anthropomorphic robotic designs \cite{ma2017openhand, mccann2021hand, Si2023DeltaHandsAS}, which lack a direct mapping to human motion, tendon-driven hands offer greater biomechanical accuracy and compliance. The human hand’s evolved structure is optimized for dexterous interactions, and much of the artificial world is designed to accommodate it. Ensuring a one-to-one mapping between human and robotic hands in mechanical design and control enhances teleoperation, imitation learning, and adaptive control strategies. Advances in imitation learning, particularly diffusion-based models \cite{Chi2023DiffusionPV, Reuss2023GoalConditionedIL, Wang2022DiffusionPA}, enable more efficient policy learning compared to traditional reinforcement learning \cite{Ahn2019ROBELRB, Andrychowicz2018LearningDI}, allowing robotic hands to perform nuanced, real-time interactions.
\TightenPar{1}

sEMG-based sensing eliminates the limitations of vision-based tracking by directly capturing the electrical potentials generated by muscle activations \cite{stashuk2001emg, merletti2016surface}. Rather than mapping sEMG signals directly to joint angles, an alternative and more robust approach is to infer movement through a musculoskeletal hand model that operates based on tendon-driven actuation \cite{Caggiano2023MyoDexAG}. This method ensures biologically consistent pose estimation and provides a stable framework for learning and predicting control signals. By integrating sEMG signals with tendon-driven control, robotic hands can achieve a level of dexterity and responsiveness closer to that of the human hand, offering new possibilities for prosthetics, rehabilitation, and teleoperated systems.

In this work, our objective is to advance research on EMG-to-tendon control models for anthropomorphic hand control by extending the existing \textit{\textit{emg2pose}} dataset. Our contributions are as follows:

\begin{itemize}
    \item \textbf{First large-scale dataset:} We introduce the first large-scale dataset specifically designed for EMG-to-tendon control signals, extending the \textit{\textit{emg2pose}} dataset with recordings from 193 subjects, 370 hours of data, and 29 stages of diverse gestures.

    \item \textbf{Comprehensive benchmarking:} We provide an extensive benchmarking framework to learn EMG-to-tendon control, tendon control-to-pose and direct EMG-to-pose mappings. \TightenPar{1}

    \item \textbf{Conditional Latent Diffusion Models (CLDM):} We present a novel approach using conditional latent diffusion model to achieve highly accurate musculoskeletal hand control signal prediction from sEMG recordings.
\end{itemize}

The paper is organized as follows: Section~\ref{sec:related} discusses related work, covering sensors (optical and sEMG) for pose tracking, tendon-based actuation, and control policy learning methods, including diffusion models and reinforcement learning. Section~\ref{sec:background} provides background on musculoskeletal hand models and sEMG, establishing the relationship between sEMG data and muscle movements. Section~\ref{sec:dataset} details the extension of the \textit{emg2pose} dataset to \textit{emg2tendon}. Section~\ref{sec:experiments}  introduces the benchmark experiments and explains the conditional latent diffusion model for modeling the EMG-to-tendon relationship. Section~\ref{sec:results} presents an extensive ablation study comparing EMG-to-tendon-to-pose and direct EMG-to-pose approaches. Section~\ref{sec:limit} discusses the limitations of the proposed dataset and methods. Finally, Section~\ref{sec:conclusion} concludes the paper and discusses future research directions. \TightenPar{1}
\section{Related Work}
\label{sec:related}

\subsection{Pose from Vision}
Vision-based pose estimation primarily uses depth, RGB, or both as inputs alongside large open-sourced datasets \cite{Mueller2017RealTimeHT,GANeratedHands_CVPR2018,Spurr2018CrossModalDV,Spurr2020WeaklyS3,spurr2021,wan2019cvpr,Boukhayma20193DHS}. Labeling approaches typically rely on marker-based motion capture systems \cite{Fan2022ARCTICAD}, which provide high-quality labels but create distributional shifts during deployment due to the absence of markers in real-world applications. Alternate labeling methods, such as multi-view camera systems \cite{Zimmermann2019FreiHANDAD,Moon2023ADO}, synthetic datasets \cite{Zimmermann2017LearningTE}, or magnetic sensors \cite{Yuan2017BigHand22MBH}, often result in lower-quality labels. While motion capture markers are well-suited for obtaining high-quality labels for surface electromyography (sEMG), they do not interfere with the data used for predictions. Vision-based datasets largely focus on static hand poses, interactions with objects, or hand-to-hand interactions.

\subsection{Pose from sEMG}
Several studies have investigated hand pose regression using sEMG signals. Liu et al. 
\cite{liu2021neuropose}, for example, used the MyoBand to estimate hand poses during various movements, collecting data from 11 participants. Their study assessed sEMG decoding models for hand pose across users and sessions using convolutional \cite{liu2021neuropose} and LSTM architectures. Similarly, SensingDynamics \cite{simpetru2022} utilized a clinic-grade system to collect datasets from 13 participants, employing custom 3D convolutional architectures to predict hand joint angles, landmark positions, and grip forces. Both studies reported low error rates on held-out test sets within participants. However, these datasets are limited in scale, featuring only 11 to 13 participants with short recording durations of 15 to 20 minutes per participant, which restricts generalization across users. In contrast, the \textit{emg2pose} dataset \cite{Salter2024emg2poseAL}, collected using the sEMG-RD wristband from CTRL-labs at Reality Labs et al. \cite{meta2024emgrd}, stands as the largest dataset available, providing a comprehensive one-to-one mapping between sEMG recordings and hand joint poses.

\subsection{Tendon based Actuation}
Musculoskeletal models \cite{mcfarland2022musculoskeletal,lee2015finger,saul2015benchmarking,delp2007opensim,seth2018opensim} have been developed to simulate the kinematics of muscles and physiological joints. While these models provide valuable insights, their high computational demands and limited support for contact forces restrict their utility for studying complex hand-object interactions, confining them to optimization-based methods. Recently, the MyoHand model \cite{caggiano2022myosuite,wang2022myosim} has addressed these limitations. MyoHand supports contact-rich interactions and is computationally efficient for data-driven exploration. Several methods have been developed for MyoHand to perform in-hand manipulation of simple objects such as balls, pens, and cubes \cite{caggiano2022myosuite,Caggiano2023MyoDexAG}, demonstrating its capability for tendon-driven manipulation tasks.

\subsection{Control Policies}
Data-driven control approaches, particularly Reinforcement Learning (RL), have demonstrated success in solving complex dexterous manipulation tasks using joint-based control \cite{rajeswaran2017learning,kumar2016manipulators,nagabandi2020deep,chen2022system}. RL techniques have been applied to enable naturalistic movements, often utilizing motion capture data to learn complex behaviors \cite{merel2017learning,merel2018neural,hasenclever2020comic}. For biomechanical models, control policies have been adapted to muscle actuators to produce more naturalistic behaviors. Unlike robotic joint-based control, biomechanical systems face challenges due to their over-actuated control spaces, which hinder exploration efficiency \cite{schumacher2022dep}. Machine learning techniques, including direct optimization \cite{pruszynski2020,ruckert2013learned} and deep reinforcement learning \cite{joos2020reinforcement,schumacher2022dep,ikkala2022breathing,caggiano2022myosuite,wang2022myosim,Caggiano2023MyoDexAG}, have been used to synthesize behaviors such as walking, running, reaching movements, and biped locomotion. While significant progress has been made in in-hand object rotations and stylistic movements \cite{Caggiano2023MyoDexAG,Berg2023SARGO}, complex dexterous hand-object manipulations remain a challenging frontier in this domain. \TightenPar{1}
\section{Background}
\label{sec:background}

We begin with a brief overview of: (i) the physiologically accurate musculoskeletal hand model (ii) the neuro-muscular interactions during finger muscle activations and how it manifests as EMG sensor data.

\subsection{Musculoskeletal Hand Model}

\begin{figure}[ht]
    \centering
    \includegraphics[width=\linewidth]{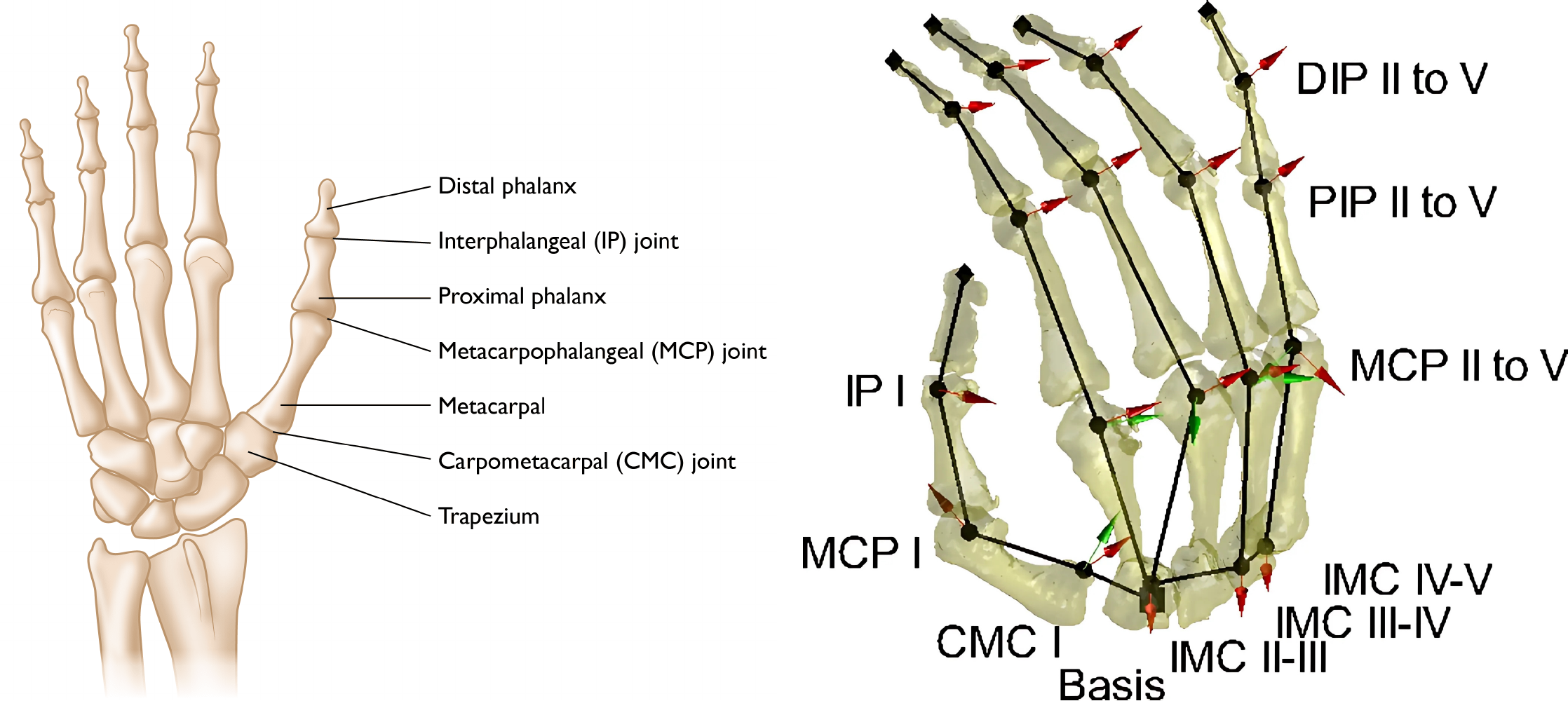}
    \caption{Anatomical structure of the human hand, showing key joints.}
    \label{fig:human_hand}
\end{figure}

The human hand consists of four fingers and a thumb, each with multiple joints that enable fine motor control and dexterous manipulation. \autoref{fig:human_hand} illustrates the skeletal structure of the human hand, highlighting key joints responsible for complex articulation. The four fingers comprise three primary joints: the metacarpophalangeal (MCP), proximal interphalangeal (PIP), and distal interphalangeal (DIP) joints. The PIP and DIP joints each have a single degree of freedom (DoF), allowing flexion and extension, while the MCP joint possesses two DoFs: flexion/extension and adduction/abduction. This results in a total of four DoFs per finger.

The thumb has a distinct anatomical structure compared to the fingers. It includes an interphalangeal (IP) joint with a single DoF for flexion/extension, while its MCP and trapeziometacarpal (TM) joints allow for both flexion/extension and adduction/abduction. Consequently, the thumb has five DoFs, making it essential for opposable grasping. Additionally, wrist motion contributes six DoFs, encompassing translation and rotation. The overall range of motion varies, with adduction/abduction movements having a more limited range compared to flexion/extension.

\begin{figure}[ht]
    \centering
    \includegraphics[width=\linewidth]{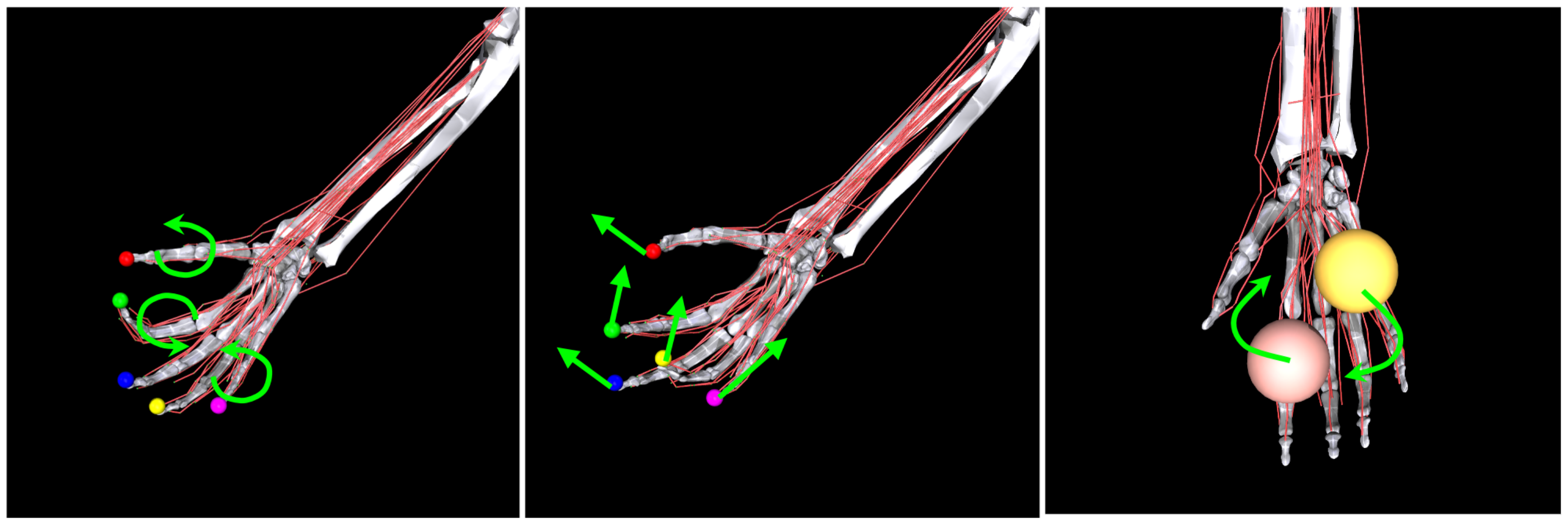}
    \caption{The MyoHand model in MyoSuite. From left to right, green arrows indicate joint articulation, fingertip movement, and in-hand manipulation of baoding balls, showcasing the model's tendon-driven control and dexterous capabilities.}
    \label{fig:myohand}
\end{figure}

To replicate human hand motion in robotic and prosthetic applications, musculoskeletal models such as MyoHand have been developed. \autoref{fig:myohand} shows the MyoSuite MyoHand model, which accurately represents human hand biomechanics, including tendon-driven actuation. Unlike direct joint-controlled robotic hands, MyoHand employs a tendon-based system, where tendons transmit forces to generate joint movements, mimicking biological muscle-tendon dynamics. This allows for greater compliance and adaptability in interaction with objects. \TightenPar{1}

To ensure biomechanical accuracy, MyoHand incorporates constraints derived from human hand kinematics. For example, normal flexion of the DIP joint is often coupled with PIP joint movement, and the MCP joint significantly limits the PIP range of motion. Such constraints are naturally captured in tendon-driven models, making them more effective in replicating human dexterity than traditional joint-based robotic hands. By leveraging inverse dynamics, we use MyoHand to generate tendon control signals corresponding to observed hand poses, providing a biologically plausible pathway for translating sEMG signals into tendon-driven robotic hand movements. \TightenPar{1}

\subsection{EMG Sensor Model}

Electromyography (EMG) sensors measure the electrical potentials generated by skeletal muscles during neurological activation. These signals provide essential insights into the temporal dynamics and morphological behavior of motor units responsible for muscle movement \cite{stashuk2001emg}. EMG signals play a crucial role in detecting and predicting muscle-driven body movements and are widely used for diagnosing neuromuscular disorders, as well as studying the physiological differences in healthy, aging, or fatigued neuromuscular systems.

To understand the muscular involvement in finger movements, we briefly outline the key muscles responsible for hand articulation. The \textit{Extensor Pollicis Longus} facilitates thumb extension, while the \textit{Abductor Pollicis Proprius} is involved in extending the index finger. The \textit{Extensor Digitorum} controls extension of the four medial fingers, and the \textit{Extensor Digiti Minimi} specifically targets the little finger. Finger adduction and abduction—movements toward or away from the midline—are governed by the \textit{Volar Interossei} and \textit{Dorsal Interossei}, which attach to the proximal phalanx and \textit{Extensor Digitorum}. Several additional muscles contribute to broader hand and arm movements, including the \textit{Supinator} for forearm rotation, the \textit{Anconeus} and \textit{Brachioradialis} for elbow control, and the \textit{Extensor Carpi Ulnaris}, \textit{Extensor Carpi Radialis Longus}, and \textit{Extensor Carpi Radialis Brevis} for wrist movement. In this study, we primarily focus on the muscles directly involved in fine finger control.

Some of these muscles are located near the skin surface and can be easily detected using surface EMG (sEMG), while others, such as the \textit{Extensor Indicis}, are deeper within the forearm. Previous research \cite{liu2021neuropose} has established a strong correlation between EMG signals and basic hand movements, highlighting the activation patterns of flexors and extensors. Additionally, studies have explored biological models for EMG signal generation in response to muscle activations, though a detailed discussion of these models is beyond the scope of this work.
\section{Proposed Dataset}
\label{sec:dataset}

In this section, we begin by summarizing the existing \textit{emg2pose} dataset, which serves as the foundation for our work. We then detail the process of generating \textit{emg2tendon}.

\subsection{\textit{emg2pose} Dataset: Brief Overview}

In this paper, we build upon the \textit{emg2pose} dataset introduced by Salter et al. \cite{Salter2024emg2poseAL}, the largest publicly available sEMG dataset to date. This dataset was collected using a 16-channel bipolar sEMG-RD wristband developed by CTRL-labs at Reality Labs \cite{meta2024emgrd}, alongside a 26-camera motion capture array operating at 60 Hz (Prime 13W OptiTrack). Participants wore sEMG-RD bands on both wrists and 19 motion capture markers on each hand, strategically placed at key joint locations, including the fingernail bases, between the DIP, PIP, and MCP joints, and at essential points on the thumb and dorsal hand. This setup enabled simultaneous high-resolution recording of muscle activity and precise hand motion tracking.

sEMG data were sampled at 2 kHz, while motion capture data were recorded at 60 Hz and later processed using an inverse dynamics solver to reconstruct hand joint angles. Due to simultaneous marker occlusions, 12.7\% of the frames could not be processed and were interpolated to match the 2 kHz sEMG sampling rate. Participants performed structured 45 to 120-second sessions containing 3-5 gestures in randomized order or unconstrained freeform movements, capturing a wide range of hand motions. Most participants contributed four sessions, with a small subset completing three. The \textit{emg2pose} dataset includes 193 participants, covering 370 hours, 751 sessions, and 29 distinct gesture categories. 

\subsection{Pose to Tendon Control Signals}

There are no direct pose-to-tendon or EMG-to-tendon datasets available, as capturing tendon movement in vivo is highly challenging. However, biophysical models describing the kinematics of tendon-driven finger joint motion exist in the literature \cite{wang2022myosim}. To overcome this limitation, we use the inverse dynamics of the MyoSuite hand model to compute tendon control signals corresponding to the pose data in the \textit{emg2pose} dataset.

\subsubsection{Muscle Actuator Dynamics}

\begin{figure*}[ht]
    \centering
    \includegraphics[width=\linewidth]{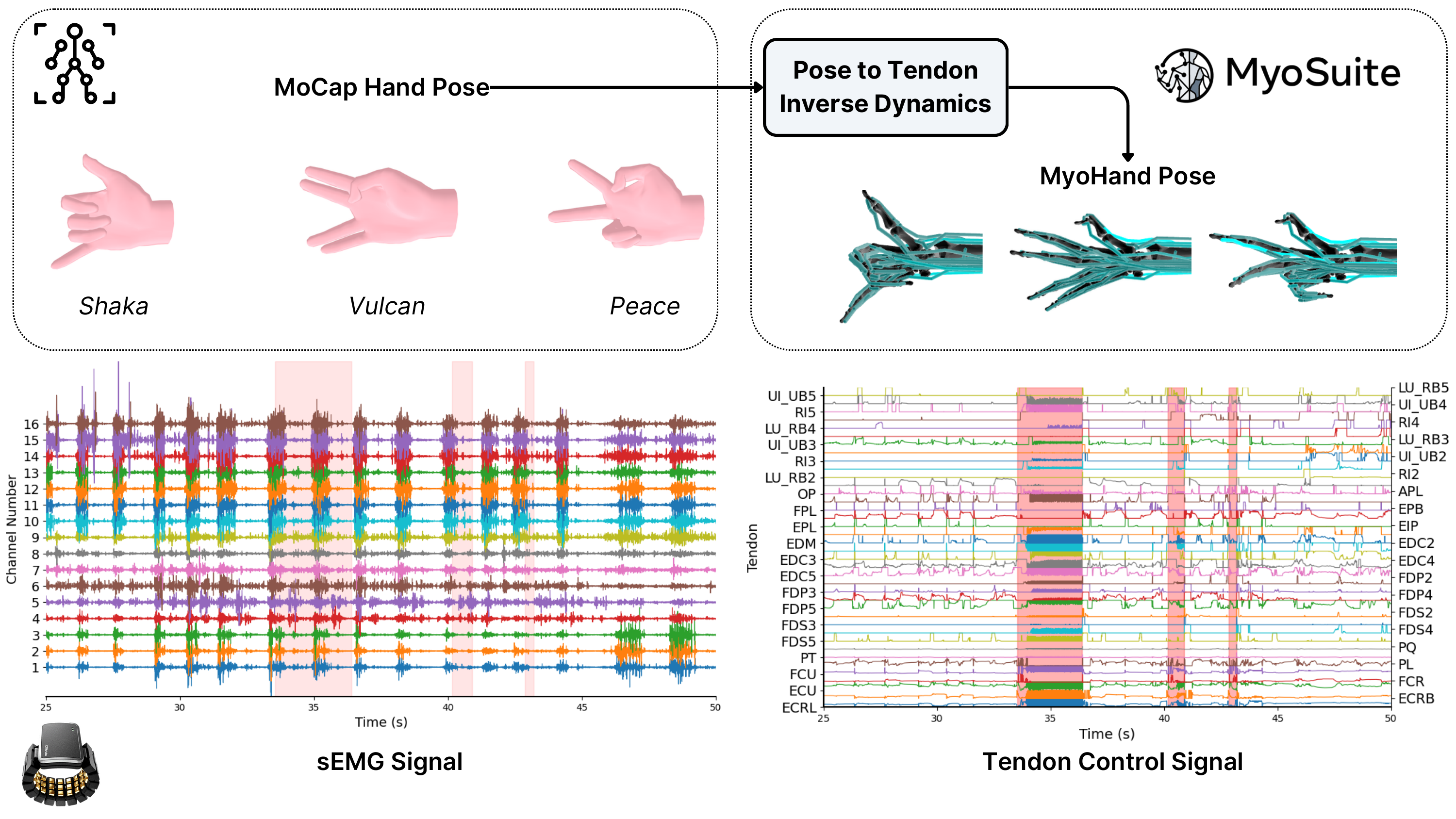}
    \caption{Overview of the \textit{emg2tendon} data collection and processing pipeline. The top-left section illustrates the \textit{emg2pose} dataset collection, where hand gestures ('Shaka,' 'Vulcan,' and 'Peace') are captured using a motion capture (MoCap) system and a wrist-worn sEMG device. The top-right section represents the MyoSuite-based musculoskeletal hand model, which enables pose-to-tendon control mapping through inverse dynamics. The bottom-left plot shows the recorded sEMG signals from 16 channels, while the bottom-right plot depicts the computed 39 channel tendon control signals.}
    \label{fig:data_gen}
\end{figure*}

In biomechanics, a muscle-tendon actuator consists of a muscle and a tendon connected in series. MuJoCo implements this by modeling the tendon as the entity with spatial properties (such as length and velocity), while the muscle is represented as a force-generating mechanism that applies tension to the tendon. Thus, in MuJoCo, the tendon length corresponds to the sum of the biological muscle and tendon lengths:

\begin{equation}
\text{actuator length} = L_T + L_M
\end{equation}

where $L_T$ is the biological tendon length, assumed inelastic, and $L_M$ is the biological muscle length, which varies dynamically. \TightenPar{1}

The optimal resting muscle length, denoted as $L_0$, is the length at which the muscle produces its peak active force at zero velocity. Due to the complexity of tendon routing and wrapping, directly specifying $L_0$ and $L_T$ is impractical. Instead, these parameters are computed automatically using predefined scaling relationships. The scaled muscle length and velocity are given by:

\begin{equation}
L = \frac{\text{actuator length} - L_T}{L_0}, \quad V = \frac{\text{actuator velocity}}{L_0}
\end{equation}

where the range constraints for muscle length, typically set as $(0.75, 1.05)$, are used to determine $L_0$ and $L_T$.

The force-length-velocity (FLV) function models the active and passive forces generated by the muscle:

\begin{equation}
FLV(L, V, act) = F_L(L) \cdot F_V(V) \cdot act + F_P(L)
\end{equation}

where:  
- $F_L(L)$ represents active force as a function of muscle length,  
- $F_V(V)$ represents active force as a function of velocity,  
- $act$ is the activation level of the muscle, and  
- $F_P(L)$ accounts for passive forces that are independent of activation.  \TightenPar{1}

The total muscle force applied to the actuator is:

\begin{equation}
\text{actuator force} = -FLV(L, V, act) \cdot F_0
\end{equation}

where $F_0$ is the peak active force at zero velocity, related to the muscle’s physiological cross-sectional area (PCSA). If unknown, $F_0$ is set to $-1$ by default and scaled using:

\begin{equation}
F_0 = \frac{\text{scale}}{\text{actuator acc}_0}
\end{equation}

where $\text{actuator acc}_0$ represents the joint acceleration caused by unit force applied to the actuator transmission.

The activation state of a muscle is governed by a first-order nonlinear filter that depends on the control signal:

\begin{equation}
\frac{d}{dt} act = \frac{ctrl - act}{\tau(ctrl, act)}
\end{equation}

where $ctrl$ represents the input control signal, clamped within $[0,1]$. The activation time constant $\tau(ctrl, act)$ is defined as:

\begin{equation}
\tau(ctrl, act) =
\begin{cases} 
\tau_{act} \cdot (0.5 + 1.5 \cdot act), & ctrl - act > 0 \\ 
\tau_{deact} \cdot (0.5 + 1.5 \cdot act), & ctrl - act \leq 0
\end{cases}
\end{equation}

where $\tau_{act}$ and $\tau_{deact}$ represent the activation and deactivation time constants, respectively. To ensure smooth transitions, MuJoCo introduces an optional smoothing parameter, $\tau_{\text{smooth}}$, which interpolates between activation and deactivation dynamics using a sigmoid function:

\begin{equation}
s(x) =
\begin{cases} 
0, & x \leq 0 \\ 
6x^5 - 15x^4 + 10x^3, & 0 < x < 1 \\ 
1, & x \geq 1
\end{cases}
\end{equation}

\subsubsection{Computation of control signals}

The control signal $ctrl$ is determined by solving the following muscle actuator equation:

\begin{equation}
AM \cdot \left( \text{gain} \odot \left( act + \text{timestep} \cdot \frac{ctrl - act}{\tau} \right) + \text{bias} \right) - q_{\text{frc}} = 0
\end{equation}

where:
- $AM$ is the actuator transmission matrix,
- $\text{gain}$ and $\text{bias}$ represent actuator gain and bias,  
- $q_{\text{frc}}$ is the joint force, and  
- $\tau$ is computed as:

\begin{equation}
\tau = \tau_D + (\tau_A - \tau_D) \cdot \text{sigmoid} \left(\frac{ctrl - act}{\tau_{\text{smooth}}} + 0.5 \right)
\end{equation}

To efficiently solve for $ctrl$, we reformulate the equation as a quadratic program (QP):

\begin{equation}
\min_{x} \frac{1}{2} x^T P x + q^T x, \quad \text{s.t.} \quad lb \leq x \leq ub
\end{equation}

where the sigmoid function is approximated to enable efficient optimization. Using this approximation, the equation is rewritten as:

\begin{equation}
AM \cdot x + k = 0
\end{equation}

where:

\begin{equation}
x = \left( \text{timestep} \cdot \text{gain} \odot \frac{ctrl - act}{(ctrl - act) \cdot \tau_1 + \tau_2} \right)
\end{equation}

\begin{equation}
k = AM \cdot (\text{gain} \odot act) + AM \cdot \text{bias} - q_{\text{frc}}
\end{equation}

From the QP formulation, we define:

\begin{equation}
P = 2 \cdot AM^T \cdot AM, \quad q = 2 \cdot AM^T \cdot k
\end{equation}

The bounds for optimization are:

\begin{equation}
lb = \text{timestep} \cdot \text{gain} \odot \frac{1 - act}{(1 - act) \cdot \tau_1 + \tau_2}
\end{equation}

\begin{equation}
ub = \text{timestep} \cdot \text{gain} \odot \frac{-act}{-act \cdot \tau_1 + \tau_2}
\end{equation}

After solving for $x$, the control signal $ctrl$ is computed as:

\begin{equation}
ctrl = act + \frac{x \cdot \tau_2}{\text{timestep} \cdot \text{gain} - x \cdot \tau_1}
\end{equation}

This framework enables us to compute tendon control signals for the \textit{emg2pose} dataset, allowing its extension into \textit{emg2tendon} for tendon-driven robotic hand control.

\subsection{Tendon Control Signals Generation}

\autoref{fig:data_gen} provides an overview of the \textit{emg2tendon} data processing workflow. The \textit{emg2pose} dataset provides hand pose recordings with 20 joints, whereas the MyoHand musculoskeletal model \cite{caggiano2022myosuite} consists of 23 joints. To ensure compatibility between these datasets, we mapped all finger joints from \textit{emg2pose} to MyoHand, covering five fingers, each with four joints. However, three wrist-related joints—pro-supination (pro\_sup), radial-ulnar deviation (deviation), and flexion-extension (flexion)—are excluded, as they are not explicitly modeled in \textit{emg2pose}.

To compute tendon control signals, we applied inverse dynamics to all pose recordings in \textit{emg2pose} using the MyoHand model, yielding 39 tendon control signals. However, in 21 sessions, the IK solver failed due to multiple occlusions in the pose data, resulting in infeasible joint configurations for MyoHand. To expedite these computations, we leveraged the MyoSuite MyoHand model, which we observed to run approximately five times faster on Apple Silicon compared to x86\_64 or ARM-based systems. Using a distributed setup of 15 Mac Mini devices (4 × M1, 6 × M2, 5 × M3), we processed inverse dynamics across all 25,254 sessions, completing the full dataset transformation in 46 hours.

To align the data format with MyoHand’s simulation requirements, we downsampled the pose recordings from 2 kHz to 500 Hz before running the IK solver. The computed tendon control signals were then upsampled back to 2 kHz to match the original \textit{emg2pose} dataset, ensuring that sEMG signals, joint poses, and tendon control forces remain synchronized at a uniform sampling rate. This pipeline facilitates seamless model training and analysis by providing a structured approach to tendon-driven hand motion estimation.
\section{Experiments}
\label{sec:experiments}

\begin{figure*}[ht]
    \centering
    \includegraphics[width=\linewidth]{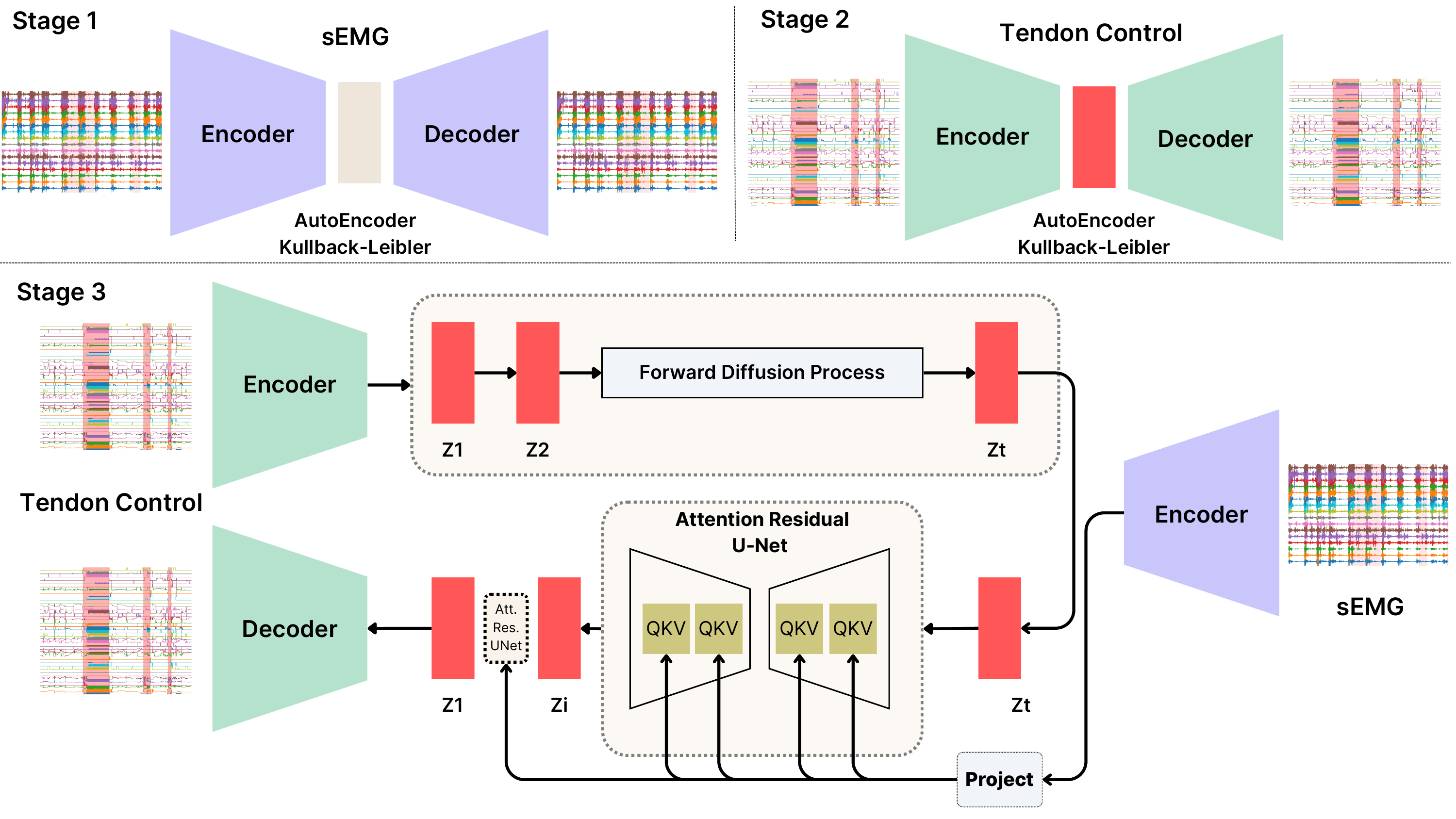}
    \caption{Training stages of Conditional Latent Diffusion Model (CLDM) to predict tendon control signals from sEMG recordings.}
    \label{fig:cldm}
\end{figure*}
\thispagestyle{empty}
\pagestyle{empty}

In this section, we demonstrate the applicability of the proposed \textit{emg2tendon} dataset by training regression models to predict tendon control signals from sEMG recordings. To evaluate the effectiveness of this approach, we compare it with EMG-to-pose regression and tendon-to-pose regression models. This enables us to assess the impact of an intermediate tendon representation on final pose estimation by analyzing the difference between direct EMG-to-pose mapping and a two-step approach (EMG $\rightarrow$ tendon $\rightarrow$ pose).

Additionally, we evaluate an alternative physics-informed setup, where predicted tendon control signals from the regression model are applied to the MyoSuite MyoHand model to generate hand pose estimations through simulation. This allows us to compare learned regression models with biomechanically accurate physics-based pose estimation.

For these experiments, we use three baseline models from \textit{emg2pose}: Time-Depth Separable Convolution (TDS) Network, NeuroPose, and SensingDynamics. We also introduce a Conditional Latent Diffusion Model (CLDM) as our proposed approach for EMG-to-tendon regression. 


\subsection{Conditional Latent Diffusion Model (CLDM)}

Several methods have been developed to apply diffusion models to time-series data, particularly in healthcare applications. We follow a similar approach, structuring the training into three main stages. First, we train two autoencoders with Kullback-Leibler (KL) regularization to compress data into a latent vector representation. One autoencoder is designed for the output signal (tendon control signals), while the other is for the input signal (sEMG recordings), which serves as the conditioning input for the diffusion model. \autoref{fig:cldm} shows the CLDM architecture including the autoencoders.

Once trained, we use the frozen encoder of the output autoencoder to extract latent vectors, which are then processed through a diffusion model. The diffusion process involves gradually adding noise to the latent representation over multiple time steps and then using a learned reverse process to reconstruct the original data. The reverse diffusion process is conditioned on the latent encoding of the input signal, ensuring that the model learns meaningful correlations between sEMG and tendon control signals. The final output is decoded using the decoder of the output autoencoder, reconstructing the predicted tendon control signals. This approach ensures that both input and output signals share a common latent space, effectively capturing the relationship between muscle activations and tendon-driven hand movements.

During inference, we generate a random noise sample from a Gaussian distribution, learned during training, and condition it using the latent vector from the input autoencoder. This conditioned latent representation is then passed through the reverse diffusion process, and the final predicted tendon control signals are obtained by decoding the output.

The autoencoder architecture consists of an Encoder and Decoder, both featuring multiple layers of ResNet Blocks, Attention Blocks, and Downsampling/Upsampling Blocks. For the diffusion process, we employ a UNet-based architecture, which effectively refines the latent representations over multiple diffusion steps.


\subsection{Networks Training}

To evaluate different approaches for predicting hand pose and tendon control signals from sEMG signals, we train multiple neural network architectures under three key tasks.

\begin{enumerate}
    \item \textbf{Pose regression from EMG:} We train Time-Depth Separable (TDS) Network, NeuroPose, and SensingDynamics to directly predict joint poses from EMG signals. Additionally, we introduce a Conditional Latent Diffusion Model (CLDM) where two autoencoders are first trained—one for joint poses and another for sEMG recordings. The CLDM is then trained to predict joint poses conditioned on the latent representation of EMG.

    \item \textbf{Tendon control signals regression from EMG:} We train the same baseline models (TDS, NeuroPose, and SensingDynamics) to predict tendon control signals from EMG recordings. Similar to pose regression, we train an autoencoder specifically for tendon control signals and then apply CLDM to model tendon control signal prediction conditioned on EMG signals.

    \item \textbf{Pose regression from tendon control signals:} Here, we train TDS, NeuroPose, and SensingDynamics to predict joint poses from tendon control signal data. Additionally, we train a CLDM model, where tendon control signals serve as the conditioning input for predicting hand pose.
\end{enumerate}


We train all three autoencoders using a 16-channel latent space with a four-layer encoder-decoder structure \([32, 64, 64, 64]\) and attention at all levels. Using these pre-trained encoders, we train three CLDMs. The CLDM architecture employs a UNet-based diffusion process with a sample size of 4000, extra conditioning channels set to 16, and downsampling across four blocks \([32, 64, 128, 256]\). The noise schedule follows a linear progression with 1000 timesteps (\(\beta_{\text{start}} = 0.0001, \beta_{\text{end}} = 0.02\)), and training is performed using the Adam optimizer (\(10^{-4}\) learning rate) with a Reduce-on-Plateau scheduler. For baseline models (TDS, NeuroPose, SensingDynamics), we adopt the same architecture configurations and training settings as in \textit{emg2pose}. All models are trained on a single NVIDIA GH200 system with a 96GB H100 GPU, 64-core ARM64 CPU, and 463.9GB RAM running Ubuntu 22.04.5. To evaluate performance, we use angular error as the primary metric for pose regression. For the prediction of the tendon control signal, we report the root mean square error (RMSE) and the mean absolute error (MAE). 
\section{Results}
\label{sec:results}

\begin{table*}[!ht]
\centering
\resizebox{\textwidth}{!}{
\begin{tabular}{l l c c c c}
\toprule
\textbf{Test Set} & \textbf{Network} & \textbf{sEMG-to-Pose} & \textbf{Tendon-to-Pose} & \textbf{Two-Step} & \textbf{Two-Step Simulation} \\
\midrule
\midrule
\multirow{4}{*}{User} 
  & SensingDynamics & 15.5 ± 1.4 & 14.6 ± 1.2 & 18.4 ± 1.8 & 16.2 ± 1.5 \\
  & NeuroPose & 13.2 ± 1.1 & 12.4 ± 1.3 & 16.7 ± 1.3 & 14.4 ± 1.1 \\
  & TDS & 12.2 ± 1.3 & 11.1 ± 1.1 & 14.5 ± 1.4 & 13.6 ± 1.3 \\
  & \textbf{CLDM} & \textbf{11.3 ± 1.0} & \textbf{10.8 ± 0.9} & \textbf{12.6 ± 1.2} & \textbf{11.9 ± 1.3} \\
\midrule
\multirow{4}{*}{Stage} 
  & SensingDynamics & 18.8 ± 1.6 & 17.6 ± 1.8 & 20.9 ± 1.5 & 19.1 ± 1.7 \\
  & NeuroPose & 17.2 ± 1.7 & 16.2 ± 1.6 & 19.8 ± 1.7 & 17.8 ± 1.5 \\
  & TDS & 15.2 ± 1.6 & 14.3 ± 1.4 & 17.9 ± 1.6 & 16.1 ± 1.6 \\
  & \textbf{CLDM} & \textbf{14.3 ± 1.5} & \textbf{12.5 ± 1.3} & \textbf{16.2 ± 1.5} & \textbf{14.9 ± 1.3} \\
\midrule
\multirow{4}{*}{User, Stage} 
  & SensingDynamics & 18.7 ± 1.6 & 17.9 ± 1.5 & 21.3 ± 1.9 & 19.2 ± 1.7 \\
  & NeuroPose & 17.5 ± 1.5 & 16.5 ± 1.7 & 21.1 ± 1.7 & 18.3 ± 1.6 \\
  & TDS & 15.8 ± 1.4 & 14.5 ± 1.6 & 18.7 ± 1.4 & 16.4 ± 1.5 \\
  & \textbf{CLDM} & \textbf{14.7 ± 1.4} & \textbf{13.1 ± 1.4} & \textbf{16.9 ± 1.3} & \textbf{15.2 ± 1.2} \\
\bottomrule
\end{tabular}}
\caption{\textnormal{Comparison of mean angular error (lower is better) for different pose estimation methods. CLDM consistently outperforms the three baselines, achieving the most accurate pose predictions. Bold values indicate statistically significant improvements.}}
\label{tab:pose_estimation_results}
\end{table*}

In this section, we present the evaluation results of our trained models for predicting both tendon control signals and joint poses from sEMG signals. We analyze the performance of different regression approaches and assess the impact of using tendon control signals as an intermediate representation for pose estimation.

We begin by reporting results for direct EMG-to-pose regression, followed by EMG-to-tendon control signal regression and tendon control signal-to-pose regression. Next, we evaluate the two-step regression approach (EMG → tendon control signals → pose) to determine whether tendon control signals provide a meaningful intermediary representation for improved pose estimation. Finally, we compare this method with a physics-informed approach, where the predicted tendon control signals are applied to the MyoSuite MyoHand model to generate pose estimations through simulation.

Table \ref{tab:pose_estimation_results} presents the pose estimation results for different models trained to predict joint angles using five different approaches: 1) pose regression from sEMG signals (direct EMG-to-Pose mapping), 2) pose regression from tendon control signals (tendon-to-pose mapping), 3) two-step pose estimation: predicting tendon control signals from EMG, followed by joint pose estimation, and 4) two-step pose estimation with simulation: using MyoSuite’s physics-based model to infer joint poses from predicted tendon control signals. \TightenPar{1}

The table reports mean angular error ± standard deviation across users, computed in the same manner as \textit{emg2pose}. Bold values indicate statistical significance under a Wilcoxon signed-rank test, comparing \textit{emg2pose} against NeuroPose and SensingDynamics.\TightenPar{1}

\begin{table}[ht]
\centering
\resizebox{0.45\textwidth}{!}{
\begin{tabular}{l l c c}
\toprule
\textbf{Test Set} & \textbf{Network} & \textbf{RMSE} & \textbf{MAE} \\
\midrule
\midrule
\multirow{4}{*}{User} 
  & SensingDynamics & 0.243 & 0.169 \\
  & NeuroPose & 0.226 & 0.155 \\
  & TDS & 0.215 & 0.150 \\
  & \textbf{CLDM} & \textbf{0.201} & \textbf{0.139} \\
\midrule
\multirow{4}{*}{Stage} 
  & SensingDynamics & 0.281 & 0.182 \\
  & NeuroPose & 0.274 & 0.163 \\
  & TDS & 0.263 & 0.157 \\
  & \textbf{CLDM} & \textbf{0.249} & \textbf{0.151} \\
\midrule
\multirow{4}{*}{User, Stage} 
  & SensingDynamics & 0.285 & 0.187 \\
  & NeuroPose & 0.281 & 0.171 \\
  & TDS & 0.269 & 0.165 \\
  & \textbf{CLDM} & \textbf{0.253} & \textbf{0.159} \\
\bottomrule
\end{tabular}}
\caption{\textnormal{Comparison of RMSE and MAE for tendon force prediction models. CLDM achieves the lowest errors, demonstrating superior accuracy compared to the three baselines. Bold values indicate statistically significant improvements.}}
\label{tab:tendon_force_results}
\end{table}

Table \ref{tab:tendon_force_results} shows Root Mean Squared Error (RMSE) and Mean Absolute Error (MAE) for networks trained to predict tendon control signals from sEMG recordings. These models are critical in both two-step regression variants, as they enable tendon control signal estimation from EMG before final pose inference. \TightenPar{1}

\subsection{Pose Regression from sEMG Signals}
Directly predicting hand pose from sEMG signals shows CLDM achieving the lowest angular error (11.3° ± 1.0°) across users, significantly outperforming all baseline models. Compared to TDS (12.2° ± 1.3°) and NeuroPose (13.2° ± 1.1°), the diffusion-based model benefits from better temporal smoothing and uncertainty modeling, reducing errors by 7.4\% over TDS and 14.3\% over NeuroPose.

\subsection{Tendon Control Signals Regression from sEMG Signals}
CLDM also achieves the best RMSE (0.201) and MAE (0.139) for tendon control signal regression, outperforming NeuroPose (0.226 RMSE, 0.155 MAE) and TDS (0.215 RMSE, 0.150 MAE). This suggests that CLDM captures fine-grained muscle activation patterns more effectively, leading to better tendon control signal predictions that enhance two-step pose estimation models.

\subsection{Pose Regression from Tendon Control Signals}
When predicting joint angles from tendon control signals, CLDM again outperforms baselines with an angular error of 10.8° ± 0.9°, demonstrating that it learns a more accurate tendon-to-pose mapping than TDS (11.1° ± 1.1°) and NeuroPose (12.4° ± 1.3°).

\subsection{Two-Step Pose Estimation: EMG → Tendon Control Signals → Pose}
Using tendon control signals as an intermediary, CLDM achieves 12.6° ± 1.2°, which is lower than direct EMG-to-Pose regression using TDS (14.5° ± 1.4°). This suggests that tendon control signals act as a meaningful latent representation, improving pose estimation.

\subsection{Two-Step Pose Estimation with Physics-Based Simulation: EMG → Tendon Control Signals → Simulated Pose}
Incorporating physics-based constraints in MyoSuite, CLDM achieves the best results (11.9° ± 1.3°), which is a 5.6\% improvement over TDS (13.6° ± 1.3°) and a 7.6\% improvement over NeuroPose (14.4° ± 1.1°). This demonstrates the importance of integrating learned tendon control signals with biomechanical simulation for achieving realistic and accurate pose estimation.

The results demonstrate that CLDM consistently outperforms all baseline models across all five evaluation settings. Direct EMG-to-Pose regression benefits from the diffusion process’s ability to capture temporal dependencies, leading to lower angular error compared to conventional deep learning architectures. Using tendon control signals as an intermediary representation further improves pose estimation accuracy, highlighting the importance of tendon-driven modeling for robotic hand control. The two-step regression approach (EMG → tendon control signals → pose) shows that incorporating tendon control as an intermediate layer enhances precision over direct pose regression. Furthermore, integrating biomechanical constraints through MyoSuite's physics-based simulation significantly improves final pose estimation, reducing errors caused by overfitting and unrealistic joint kinematics. CLDM-based models achieve the lowest errors across all tasks, underscoring their effectiveness in learning the complex relationships between muscle activations and tendon control signals. These findings indicate that the use of latent diffusion and tendon-based modeling can drive substantial progress in tendon-driven robotic hand control, prosthetics, and teleoperation applications.

\section{Limitations}
\label{sec:limit}

Through our experiments, we identified two main limitations that highlight potential areas for improvement in our approach. First, our method heavily relies on accurate pose estimation and inverse dynamics for mapping joint poses to tendon control signals. The \textit{emg2pose} dataset already exhibits a 12.7\% failure rate in its inverse dynamics solver due to marker occlusions and infeasible joint positions. Additionally, failures in the pose-to-tendon mapping further compound the issue. These limitations stem primarily from data collection constraints rather than the model itself.

Second, the process of obtaining tendon control signals from joint poses via inverse dynamics applied to a simulated musculoskeletal hand model inherently places the dataset within a synthetic space. While the MyoHand model serves as a useful approximation, a definitive evaluation of our system requires real-world testing on physical tendon-driven robotic hands. The current MyoHand model has been simplified to balance computational efficiency and interpretability, but it omits several secondary muscles that contribute to the full range of motion of the human hand. This simplification may introduce discrepancies between the predicted tendon forces and those observed in actual biomechanical systems.
\section{Conclusion} 
\label{sec:conclusion}

In this work, we extend the \textit{emg2pose} dataset to create the first large-scale dataset for mapping sEMG signals to tendon control signals, \textit{emg2tendon}. We establish three baseline models on this dataset and evaluate their performance in predicting tendon control signals. Additionally, we introduce a diffusion-based method that demonstrates strong performance not only for \textit{emg2pose} but also for the proposed \textit{emg2tendon} mapping problem, highlighting its potential for accurate and robust tendon-driven hand control.

Moving forward, we aim to validate our approach on real tendon-driven robotic hands to bridge the gap between synthetic and physical systems. Additionally, we seek to explore reinforcement learning (RL) techniques to enhance the performance of diffusion-based models for tendon control signal prediction. Lastly, we plan to extend the dataset to include interactions between hands and objects, further improving its applicability to real-world dexterous manipulation tasks.



\bibliographystyle{plainnat}
\bibliography{references}

\end{document}